\documentclass[10pt,twocolumn,letterpaper]{article}

\usepackage{cvpr}
\usepackage{times}
\usepackage{epsfig}
\usepackage{graphicx}
\usepackage{amsmath}
\usepackage{amssymb}
\usepackage{multirow}

\usepackage[pagebackref=true,breaklinks=true,letterpaper=true,colorlinks,bookmarks=false]{hyperref}

\cvprfinalcopy 

\def\cvprPaperID{6082} 

\ifcvprfinal\fi
\begin{document}

\title{Analyzing the Dependency of ConvNets on Spatial Information }

\author{Yue Fan \qquad Yongqin Xian \qquad Max Maria Losch \qquad Bernt Schiele\\
Max Planck Institute for Informatics\\
Saarland Informatics Campus\\
{\tt\small yfan@mpi-inf.mpg.de \qquad yxian@mpi-inf.mpg.de \qquad mlosch@mpi-inf.mpg.de \qquad schiele@mpi-inf.mpg.de}
}

\maketitle

\begin{abstract}
Intuitively, image classification should profit from using spatial information. Recent work, however, suggests that this might be overrated in standard CNNs. In this paper, we are pushing the envelope and aim to further investigate the reliance on spatial information. We propose spatial shuffling and GAP+FC to destroy spatial information during both training and testing phases. Interestingly, we observe that spatial information can be deleted from later layers with small performance drops, which indicates spatial information at later layers is not necessary for good performance. For example, test accuracy of VGG-16 only drops by 0.03\% and 2.66\% with spatial information completely removed from the last 30\% and 53\% layers on CIFAR100, respectively. Evaluation on several object recognition datasets~(CIFAR100, Small-ImageNet, ImageNet) with a wide range of CNN architectures~(VGG16, ResNet50, ResNet152) shows an overall consistent pattern. 
\end{abstract}

\section{Introduction}

Despite the impressive performances of convolutional neural networks (CNNs) on computer vision tasks~\cite{long2015fully, girshick2014rich, alexnet2012, he2016deep, simonyan2014very}, their inner workings remain mostly obfuscated to us and analyzing them often results in surprising observations~\cite{zhang17understanding, s.2018on, brendel2019approximating, geirhos2018imagenet, ilyas2019adversarial, adver2014first}.

Generally, the majority of modern CNNs for image classification learn spatial information across all the convolutional layers: every layer in AlexNet, VGG, Inception, and ResNet applies $3\times 3$ or larger filters. 
Such design choices are based on the assumption that spatial information remains important at every convolutional layer to consecutively increase the access to a larger spatial context.
This is based on the observations that single local features can be ambiguous and should be related to other features in the same scene to make accurate predictions~\cite{torralba2003context, hoiem2008putting}.

However, recent works on restricting the receptive field of CNN architectures for scrambled inputs~\cite{brendel2019approximating} or using wavelet feature networks of shallow depth \cite{oyallon2017scaling}, have all found it to be possible to acquire competitive performances on the respective tasks. 
This raises doubts on the necessity of spatial information for classification and whether the model is still able to maintain the performance if the spatial information is completely removed from the training process. 

\begin{figure}
    \centering
    \includegraphics[width=1\linewidth]{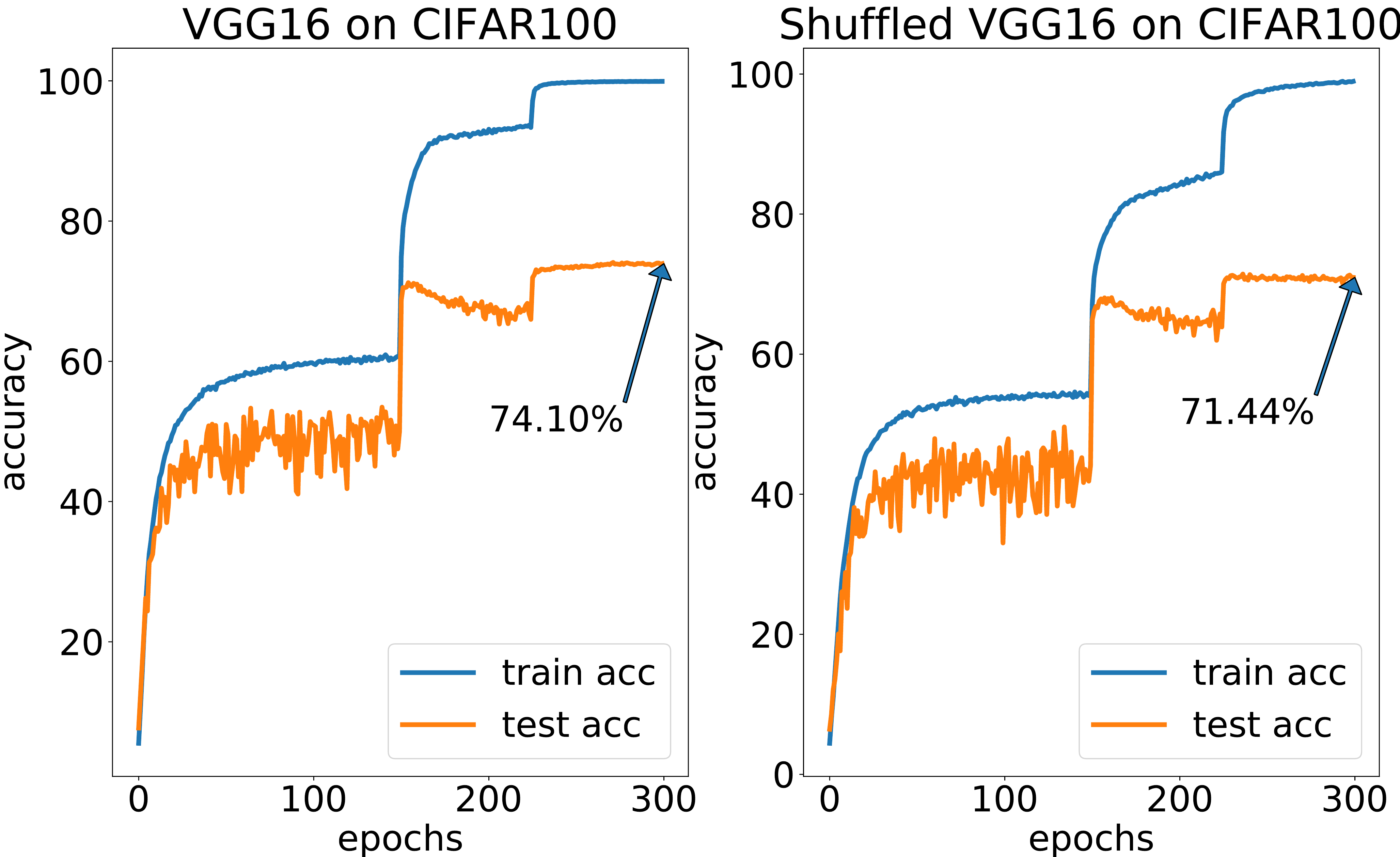}
    \caption{
    Training processes of a standard VGG-16 and a shuffled VGG-16 where feature maps in the last 54\% layers are randomly and spatially shuffled at both training and testing phases. The final test accuracy only drops 2.66\% (from 74.10\% to 71.44\%) and the training curves are very similar, which implies that spatial information maybe not necessary for a good classification accuracy.
    }
    \label{fig:train_curve}
\end{figure}
We add to the list of surprising findings surrounding the inner workings of CNNs and present a rigorous investigation on the necessity of spatial information in standard CNNs by avoiding learning spatial information at multiple layers. Spatial information refers to the spatial ordering on the feature map.

To this end, we propose \textit{channel-wise shuffle} to eliminate channel information, and \textit{spatial shuffle}, \textit{patch-wise spatial shuffle} and \textit{GAP+FC} to eliminate spatial information.
Surprisingly, we find that the modified CNNs i.e. without accessing any spatial information at later layers, can still achieve competitive results on several object recognition datasets.
For example, Fig. \ref{fig:train_curve} shows training processes of a standard VGG-16 and a shuffled VGG-16 on CIFAR100. In the shuffled VGG-16, we perform random shuffle on its feature maps spatially from the last 54\% layers at each training step. 
Interestingly, the test accuracy only drops 2.66\% and the training process is nearly identical to the standard VGG-16.
This observation generalizes to various CNN architectures: removing spatial information from the last 30\% layers gives a surprisingly little performance decrease within 1\% across architectures and datasets, and the performance decrease is still within 7\% even if the last 50\% layers are manipulated.
This indicates that spatial information is overrated for standard CNNs and not necessary to reach competitive performances.
Finally, our investigation on the detection task shows that although the unavailability of spatial information at later layers does harm the localization of the model, the impact is not as fatal as expected; at the same time, the classification ability of the model is not affected.

In our experiments, we find that spatial information at later layers is not really necessary for a good classification performance and that even though the depth of the network plays an important role, the later layers do not necessarily have to be convolutions. 
As a side effect, GAP+FC leads to a smaller model with less parameters with small performance drops.

\section{Related Work}
Training models for the task of object recognition, our intuitive understanding would be that global image context is beneficial for making accurate predictions.
For that reason extensive efforts have been made to enhance the aggregation of spatial information in the decision-making progress of CNNs.
\cite{dai2017deformable, zhu2019deformable} have made attempts to generalize the strict spatial sampling of convolutional kernels to allow for globally spread out sampling and \cite{zhao2017pyramid} have spurred a range of follow-up work on embedding global context layers with the help of spatial down-sampling.

While all of these works have improved on a related classification metric in some way, it is not entirely evident whether the architectural changes alone can be credited, as there is an increasing number of work on questioning the importance of the extent of spatial information for common CNNs.
One of the most recent observations by \cite{brendel2019approximating} for example indicates that the VGG-16 architecture trained on ImageNet is invariant to scrambled images to a large extent, e.g. they reported only a drop of slightly over $10\%$ points top-5 accuracy for a pre-trained VGG-16.
Furthermore, they construct a modified ResNet architecture with a limited receptive field as small as $33 \times 33$ and reach competitive results on ImageNet,
similar to the style of the traditional Bag-of-Visual-Words.
The latter is also explicitly incorporated into the training of CNNs in the works by \cite{mohedano2016bags, feng2017bag, cao2017local}, the effect of neglecting global spatial information by design has surprisingly little effect on performance values.
In contrast to their work, we make a clear distinction between first and last layers, and we show empirically spatial information at last layers are not necessary for good performance.

\cite{hinton17capsule} assumes that current CNNs don't respect the spatial information due to the pooling operation; CNNs look for features in the image without paying attention to their pose during prediction.
This limitation motivates the work of \cite{hinton17capsule} where they make use of dynamic routing among capsules to encode the spatial information.

Global average pooling is used to substitute the final fully connected layer in many recent works \cite{lin2013network, he2016deep}, which also hints that removing spatial information at the last layer does not affect the performance since the model is normally deep enough to obtain a sufficiently large receptive field. 

On a related note, \cite{geirhos2018imagenet} indicates that models trained solely on ImageNet do not learn shape sensitive representations with constructing object-texture mismatched images, which would be expected to require global spatial information. Instead, the models are mostly sensitive to local texture features.

Our work is motivated to push the envelope further  to investigate the necessity of spatial information in the processing pipeline of CNNs.
While the related work has put the attention mainly on altering the input and does not differentiate between last and first layers, we are interested in taking measures that remove the spatial information at different intermediate layers to shed light on how CNNs process spatial information, evaluating its importance and providing insights for architectural design choices.

\section{Methods and Experimental Setup} \label{sec:GAP+FC_1x1}



In this section, we develop methods to test how information is represented throughout the network's layers and apply these to well established architectures.
Section \ref{subsec:methods} elaborates details on our approaches and the experimental setup is discussed in section \ref{subsec:setup}.




\begin{figure}
    \centering
    \includegraphics[width=1\linewidth]{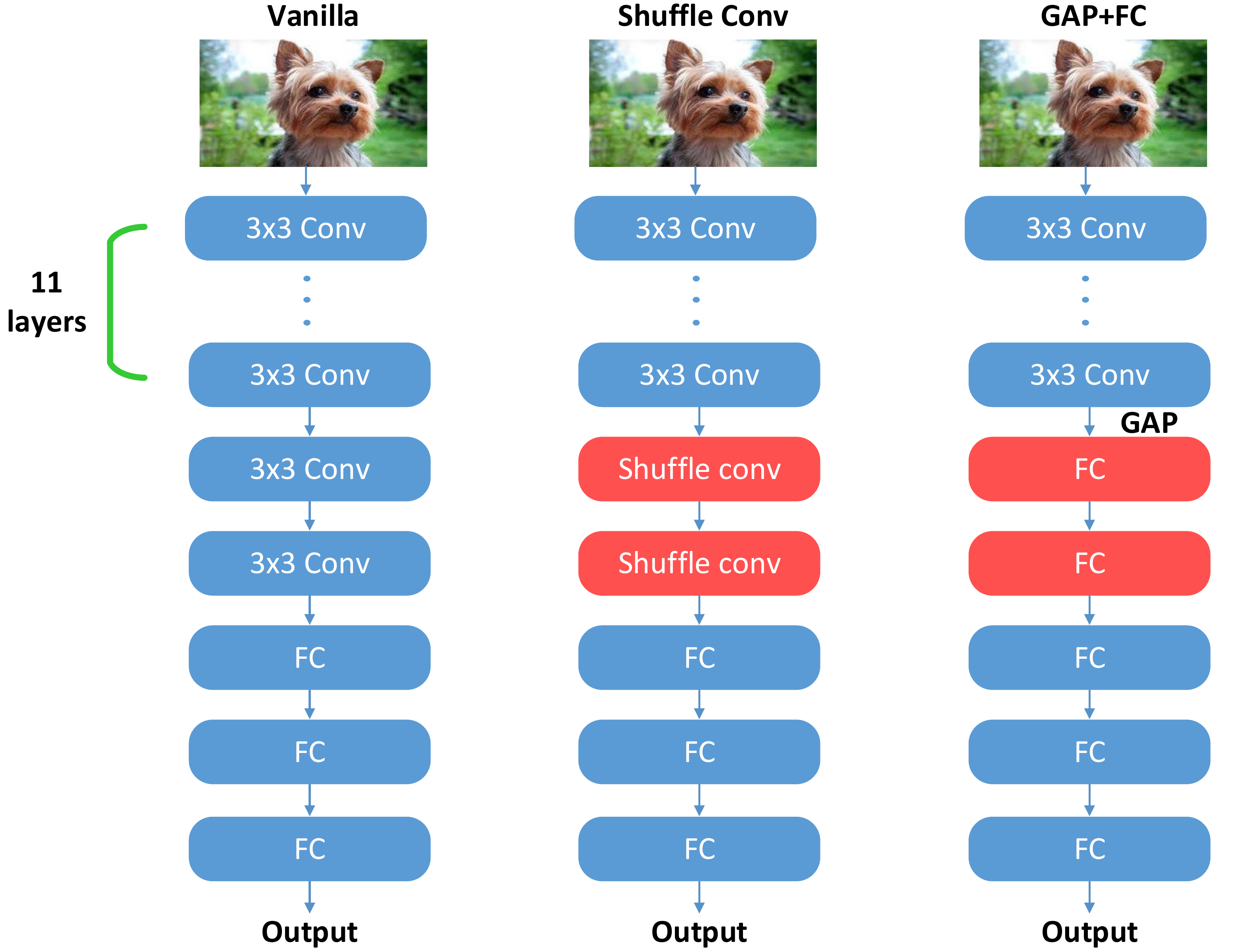}
    \caption{
    To demonstrate how we modify an existing architecture, an example of VGG-16 with the modification on its last 2 layers is shown here.
    In the middle plot, the last 2 conv layers are replaced by shuffle conv, which consists of one of the shuffling methods (\textit{channel-wise shuffle}, \textit{spatial shuffle}, \textit{patch-wise spatial shuffle}) and an ordinary convolution.
    In the left plot, the last 2 conv layers are replaced by fully-connected layers after a GAP layer.
    }
    \label{fig:arch}
\end{figure}
\begin{figure}
    \centering
    \includegraphics[width=1\linewidth]{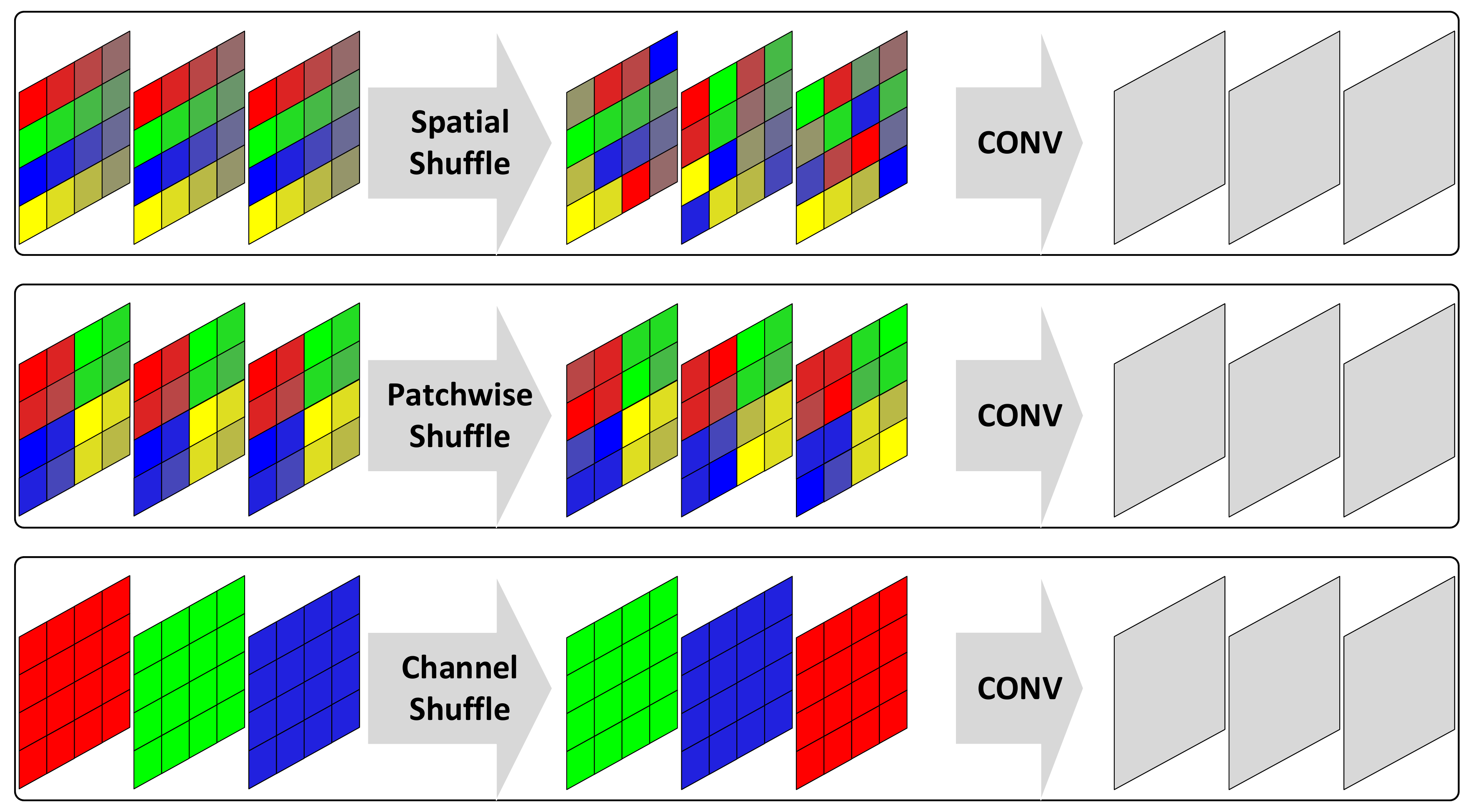}
    \caption{
    Details of 3 shuffling methods. Shuffling is always applied before the convolution.
    Spatial shuffle randomly and independently permutes pixels on each feature map at a global scale in the sense that a pixel can end up anywhere on the feature map.
    Patch-wise shuffle randomly permutes the pixel locations within a given patch and independently across patches and feature maps.
    Channel shuffle randomly permutes the order of feature maps, leaving the spatial ordering unchanged.
    }
    \label{fig:3shuffle}
\end{figure}
\subsection{Approaches to Constrain Information} \label{subsec:methods}
We propose 4 different methods, namely \textit{channel-wise shuffle}, \textit{spatial shuffle}, \textit{patch-wise spatial shuffle} and \textit{GAP+FC}, to remove either spatial or channel information from the training. 
Spatial information here refers to the awareness of the relative spatial position between activations on the feature map, and channel information stands for the dependency across feature maps.
Fig. \ref{fig:arch} middle illustrates an example of VGG-16 with its last 2 layers modified by any of the 3 shuffle method, and Fig. \ref{fig:arch} right demonstrates the same modification by GAP+FC.

\textbf{Spatial Shuffle} extends the ordinary convolution operation by prepending a random spatial shuffle operation to permute the input to the convolution. 
As illustrated in Fig. \ref{fig:3shuffle} top: Given an input tensor of size $c \times h \times w$ with $c$ being the number of feature maps for a convolutional layer, 
we first take one feature map from the input tensor and flatten it into a 1-d vector with $h \times w$ elements, whose ordering is then permuted randomly. 
The resulting vector is finally reshaped back into $h \times w$ and substitute the original feature map. 
This procedure is independently repeated $c$ times for each feature map so that activations from the same location in the previous layer are misaligned, thereby preventing the information from being encoded by the spatial arrangement of the activations.
The shuffled output becomes the input of an ordinary convolutional layer in the end. 
Even though shuffling itself is not differentiable, gradients can still be propagated through in the same way as Max Pooling. 
Therefore it can be embedded into the model directly for end-to-end training.
As the indices are recomputed within each forward pass, the shuffled output is also independent across training and testing steps. 

Images within the same batch are shuffled in the same way for the sake of simplicity since we find empirically that it does not make a difference whether the images inside the same batch are shuffled in different ways. 

\textbf{Patch-wise Spatial Shuffle}
is a variant of \textit{spatial shuffle} which is performed at a global scale in the sense that the activation can end up with an arbitrary location on the feature map.
Patch-wise spatial shuffle first divides the feature map into grids and shuffles activations within each grid independently.
Afterwards, an ordinary convolution is performed as usual.
Note that the two operations are equivalent when the patch size is the same as the feature map size.
Fig. \ref{fig:3shuffle} middle demonstrates an example of patch-wise spatial shuffle with a $2 \times 2$ patch size, where the random permutation of pixel locations is restricted within each patch.

\textbf{Channel-wise Shuffle} keeps the spatial ordering of activations and randomly permutes the ordering of feature maps to prevent the model from utilizing channel information, which is considered essential~\cite{xie2017aggregated, zhang2017interleaved, zhang2018shufflenet}.
It is used to make comparison of the model robustness against the loss of spatial information and channel information.
An illustration can be seen in Fig. \ref{fig:3shuffle}, channel-wise shuffle is also performed independently across training and testing steps.


\textbf{GAP+FC}
denotes Global Average Pooling and Fully Connected Layers.
\textit{Spatial Shuffle} is an intuitive way of destroying spatial information but it also makes it hard to learn correlations across channels for a particular spatial location.
Furthermore, shuffling introduces undesirable randomness into the model so that during evaluation multiple forward passes are needed to acquire an estimate of the mean of the output.
A simple deterministic alternative achieving a similar goal is to deploy Global Average Pooling (GAP) after an intermediate layer, and all the subsequent ones are substituted by fully connected layers. Compared to \textit{Spatial Shuffle} that introduces an extra computational burden at each forward pass, it is a much more efficient way to avoid learning spatial information at intermediate layers because it shrinks the spatial size of all subsequent feature maps to one, therefore, the number of FLOPs and parameters are also less. 


\subsection{Experimental Setup} \label{subsec:setup}

We test different architectures on 4 datasets: CIFAR100, Small-ImageNet-32x32 \cite{chrabaszcz2017downsampled}, ImageNet and Pascal VOC 2007 + 2012. Small-ImageNet-32x32 is a down-sampled version of the original ImageNet (from $256 \times 256 $ to $32 \times 32$).
We report the top-1 accuracy and mAP~\cite{voc2012, voc2007} in classification and detection experiments, respectively.  
We will take an existing model and apply the modification to different layers. The rest of the setup and hyper-parameters for modified models remain the same as the baseline models.

\textbf{Classification:}
For the VGG architecture, the modification is only performed on the convolutional layers as illustrated in Fig. \ref{fig:arch}. 
For the ResNet architecture, one bottleneck sub-module is considered as a single piece and the modification is applied onto the $3 \times 3$ convolutions within since they are the only operation with spatial extent. Features that go through the skip connection branch are also shuffled in the shuffle experiments to prevent the model from learning to ignore the information from the convolution branch. 
The rest of the configuration remains the same as in the baseline model (see supplemental material for an example of modified ResNet-50 architecture).

For CIFAR100 and Small-ImageNet-32x32 experiments, the original ResNet architecture down-samples the input image by a factor of 32 and gives $1 \times 1$ feature maps at last layers, therefore shuffling is noneffective. To make shuffling non-trivial, we set the first convolution in ResNet to $3 \times 3$ with stride $1$ and the first max pooling layer is removed so that the final feature map size is $4 \times 4$. 

In our experiments, we always first reproduce the original result on the benchmark as our baseline, and then the same training scheme~\cite{he2016deep} is directly used to train the modified models. All models in the same set of experiments are trained with the same setup from scratch and they share the same initialization from the same random seed. During testing, we make sure to use a different random seed than during training.

\textbf{Detection:}
We use training set and validation set of VOC 2012+2007 as the training data and report mAP on VOC 2007 test set. We shuffle the layer in the backbone model to test the robustness of localization against the absence of spatial information.

\section{Results} \label{sec:results}

We first compare the performance of VGG-16 on CIFAR100 with spatial or channel information missing from different number of later layers in section \ref{sec:results_shuffle}.
An in-depth study of our main observations on CIFAR100, Small-ImageNet-32x32 and ImageNet for VGG-16, ResNet-50 and ResNet-152 is conducted in section \ref{sec:results_main}.
In section \ref{sec:single}, we test whether the model performance suffers more from spatial shuffle at consecutive layers than at a single layer.
In section \ref{sec:localshuffle}, we investigate the model robustness against the loss of spatial information in various degree by controlling the amount of spatial information that can be passed through the network.
Finally, we present the result of detection on VOC datasets in section \ref{sec:det}.

\subsection{Spatial and Channel-wise Shuffle on VGG-16} \label{sec:results_shuffle}

In this section, we first investigate the invariance of pre-trained models to the absence of the spatial or channel information at test time, then we impose this invariance at training time with methods in section \ref{subsec:methods}.

\textbf{Shuffle the Last 30\% Layers Channel-wise : }
A VGG-16 trained on CIFAR100 that achieves 74.10\% test accuracy is used as our baseline. 
We first test its robustness against the absence of the channel information at test time by substituting the last 30\% convolutional layers with convolution with channel-wise shuffle.
As is expected, the test accuracy drops to 1.04\% (table \ref{tab:mismatch}) which is the same as random guess on CIFAR100.
Following the same training scheme of the baseline, we then train another VGG-16 with channel-wise shuffle added to its last 30\% convolutional layers at training time.
This model is able to reach around 67\% test accuracy no matter whether channel-wise shuffle is applied at test time.
However, it performs significantly worse than the baseline performance, which means the expressiveness of the model is much limited without utilizing the ordering of feature maps even though the spatial information is preserved. 
\begin{table}[]
\centering
\begin{tabular}{cc|c}
    \multicolumn{2}{c|}{Schemes} & \multirow{2}{*}{Top-1(\%)} \\
    Train & Test &  \\ \hline \hline
    no shuffle & no shuffle & 74.10 \\
    \hline 
    channel shuffle & channel shuffle & 67.56 \\
    channel shuffle & no shuffle & 67.80 \\
    no shuffle & channel shuffle & 1.04\\
    \hline 
    spatial shuffle & spatial shuffle & 74.07 \\
    spatial shuffle & no shuffle & 73.74 \\
    no shuffle & spatial shuffle & 23.49
\end{tabular}
\vspace{7pt}
\caption{
Top-1 accuracy of VGG-16 on CIFAR100 with spatial / channel-wise shuffle enabled at either training or test time for the last 30\% layers.
The pre-trained model is more robust to spatial shuffle (23.49\%) than channel-wise shuffle (1.04\%) at test time; when imposed in training, the model achieves 74.07\% test accuracy for spatial shuffle and 67.56\% for channel-wise shuffle, showing an impressive robustness to the loss of spatial information. 
}
\label{tab:mismatch}
\end{table}

\textbf{Shuffle the Last 30\% Layers Spatially: }
As a comparison to channel information, we repeat the same experiment on spatial shuffle and the result is presented in the second half of the table \ref{tab:mismatch}.
No shuffle$\,\to\,$spatial shuffle of the pre-trained VGG-16 gives $23.49\%$ test accuracy, which is similar to the performance of a one-hidden-layer perceptron (with 512 hidden units and ReLU activation) on CIFAR100 (25.61\%), when evaluated with random spatial shuffle.
However, if the spatial shuffle is infused into the model at training time, then the baseline performance can be retained no matter whether random spatial shuffle appears at test time (74.07\% for spatial shuffle$\,\to\,$spatial shuffle and 73.74\% for spatial shuffle$\,\to\,$no shuffle).

\begin{figure}
	\centering
	\includegraphics[width=0.9\linewidth]{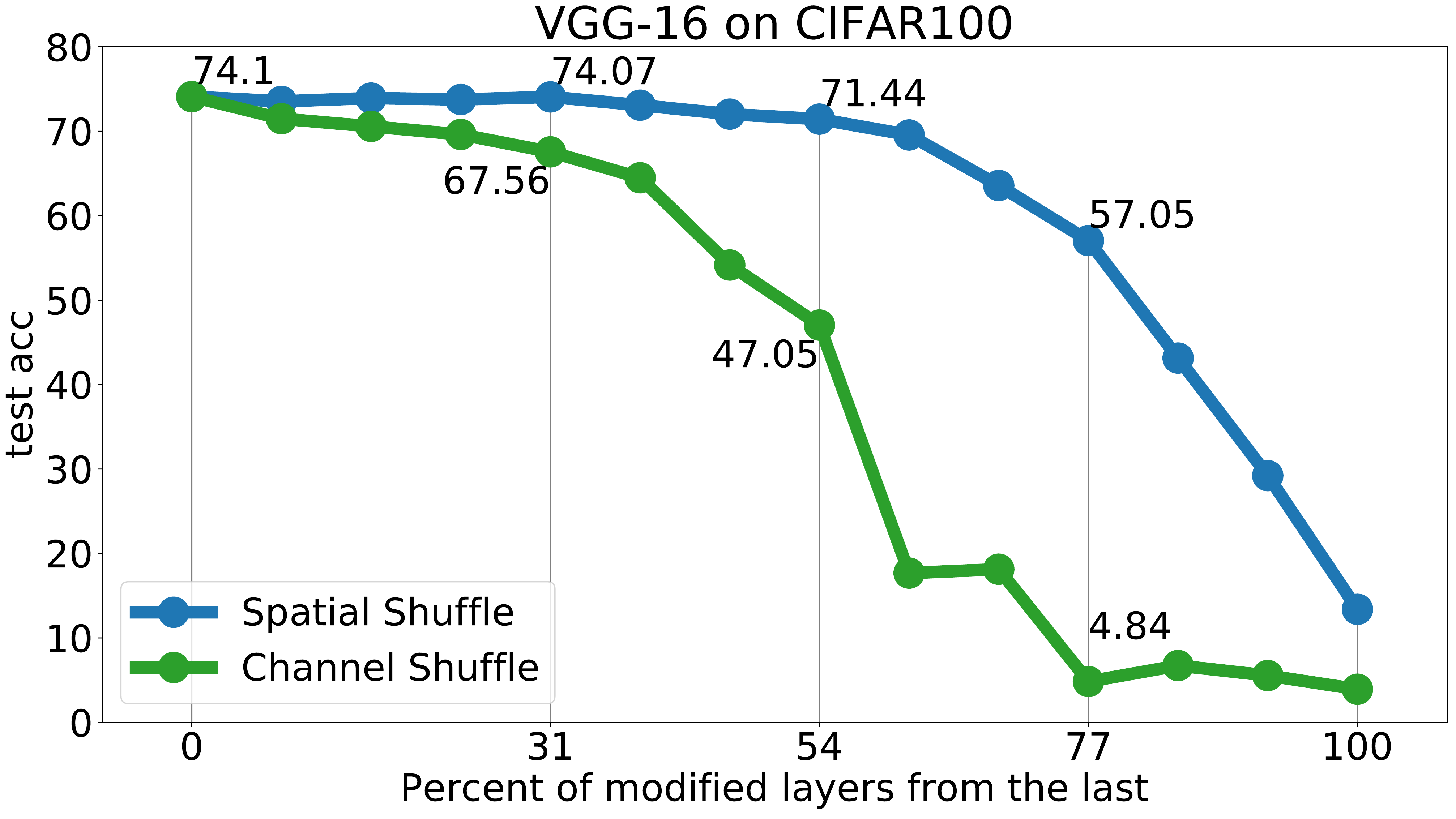}
	\caption{
	Classification accuracy of VGG-16 on CIFAR100 with different shuffle schemes.
	All models are trained with the same setup and shuffling is performed both at training and test time.
	The very slow decrease of the test accuracy of spatial shuffle implies a far less important role of spatial information for classification. 
	The test accuracy is not really affected given 31\% of its layers are modified by spatial shuffle. 
	Even with 54\% later layers being shuffled spatially the test accuracy only decreases by 2.66\%, and the same number of the performance decrease in channel-wise shuffle happens when the last layer is modified.
	}
	\label{fig:channel_shuffle}
\end{figure}

\textbf{Shuffle Other Layers: }
To systematically study the impact of spatial and channel information, we gradually increase the number of modified layers from the last in VGG-16 and report the corresponding test accuracy in Fig. \ref{fig:channel_shuffle}.
All models are trained with the same setup and shuffling is performed both at training and test time; the x-axis is the percentage of modified layers counting from the last layer on with 0 referring the baseline. 

Besides an overall decreasing trend for both shuffling with the increase of the percent of modified layers, the test accuracy of spatial shuffle drops unexpectedly slow, e.g
merely 2.66\% test accuracy drop when up to 54\% of layers from the last are shuffled spatially. 
Likewise, when spatial information is removed from the last 77\% layers it still has a reasonable performance (57.05\%), where as the performance of channel-wise shuffle is only 4.84\%.

\textbf{Discussion: }
This indicates that although a standard model makes use of both spatial dimension and channel dimension to encode information, the spatial information plays a surprisingly less pivotal role than the channel information. 
The model is even able to adapt to the complete absence of spatial information at later layers if spatial information is removed explicitly at training time, which strengthens the claims from \cite{brendel2019approximating, hinton17capsule} that CNNs intrinsically possess invariance to spatial relationship among features to some extent.
And the unsuccessful adaptation to channel-wise shuffle implies that the large model capacity may mainly come from the channel order, shuffling which causes unrecoverable damage to the model.

\subsection{Spatial Information at Later Layers are Not Necessary} \label{sec:results_main}

In this section, we design more experiments to study the reliance of different layers on spatial information: we modify the last convolutional or bottleneck layers of VGG-16 or ResNet-50 by \textit{Spatial Shuffle} (both at training and test time) and \textit{GAP+FC} such that the spatial information is removed in different ways. 
Our modification on the baseline model always starts from the last layer and is consecutively extended to the first layer. 
The modified networks are then trained on the training set with the same setup and evaluated on the hold-out validation set.

\begin{figure*}[!ht]
	\centering
	\begin{minipage}{0.35\linewidth}
	    \hspace{3cm}
		\includegraphics[scale=0.5]{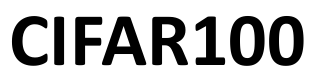}
	\end{minipage} \hspace{3cm}
	\begin{minipage}{0.35\linewidth}
	    \centering
		\includegraphics[scale=0.5]{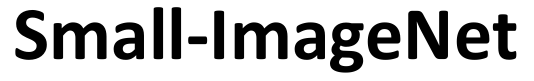}
	\end{minipage} 
	
	\begin{minipage}{0.05\linewidth}
	    \centering
		\includegraphics[scale=0.5,angle=90,origin=c]{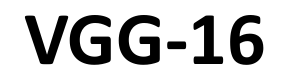}
	\end{minipage} 
	\begin{minipage}{0.45\linewidth}
	    \centering
		\includegraphics[width=1\linewidth]{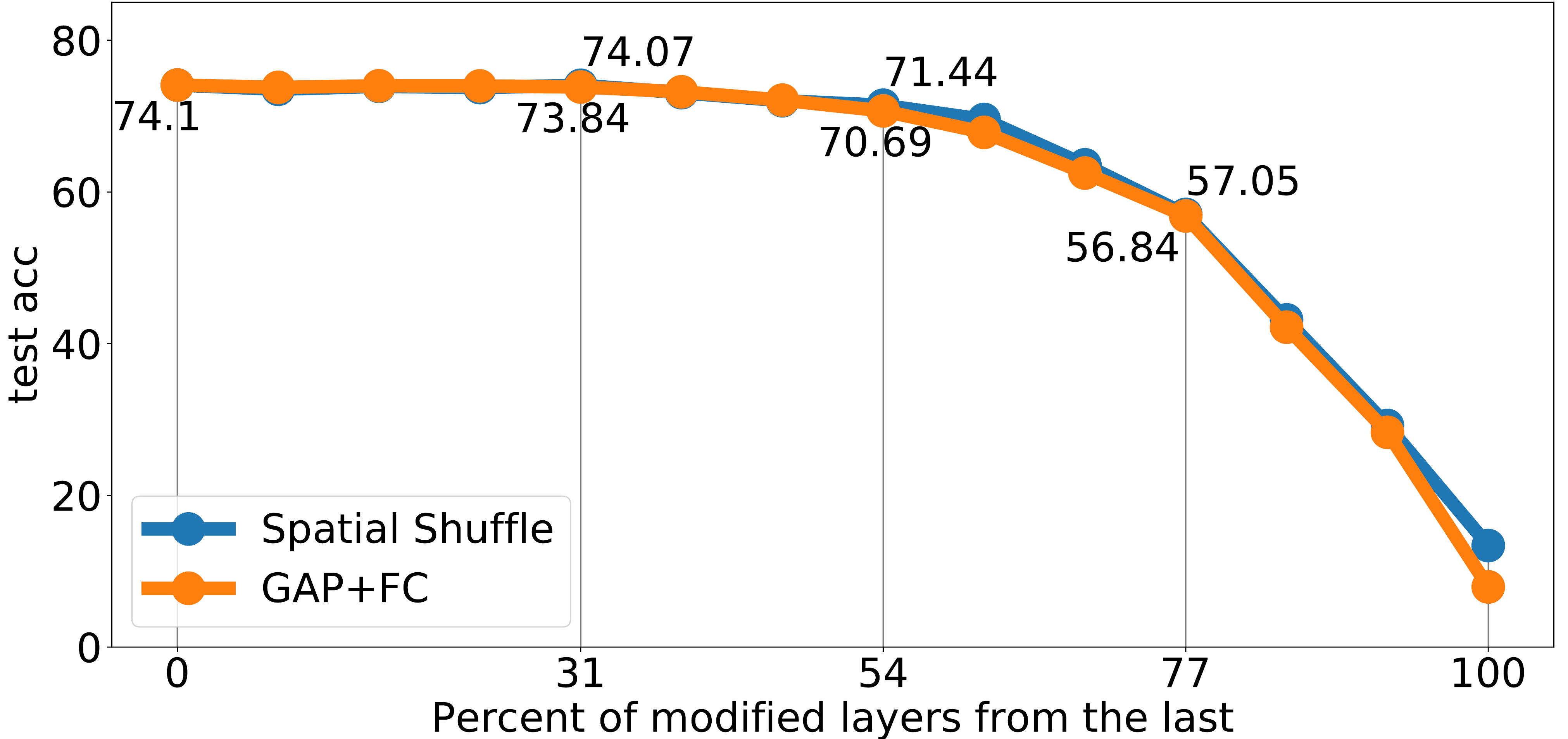}
	\end{minipage} 
	\begin{minipage}{0.45\linewidth}
	    \centering
		\includegraphics[width=1\linewidth]{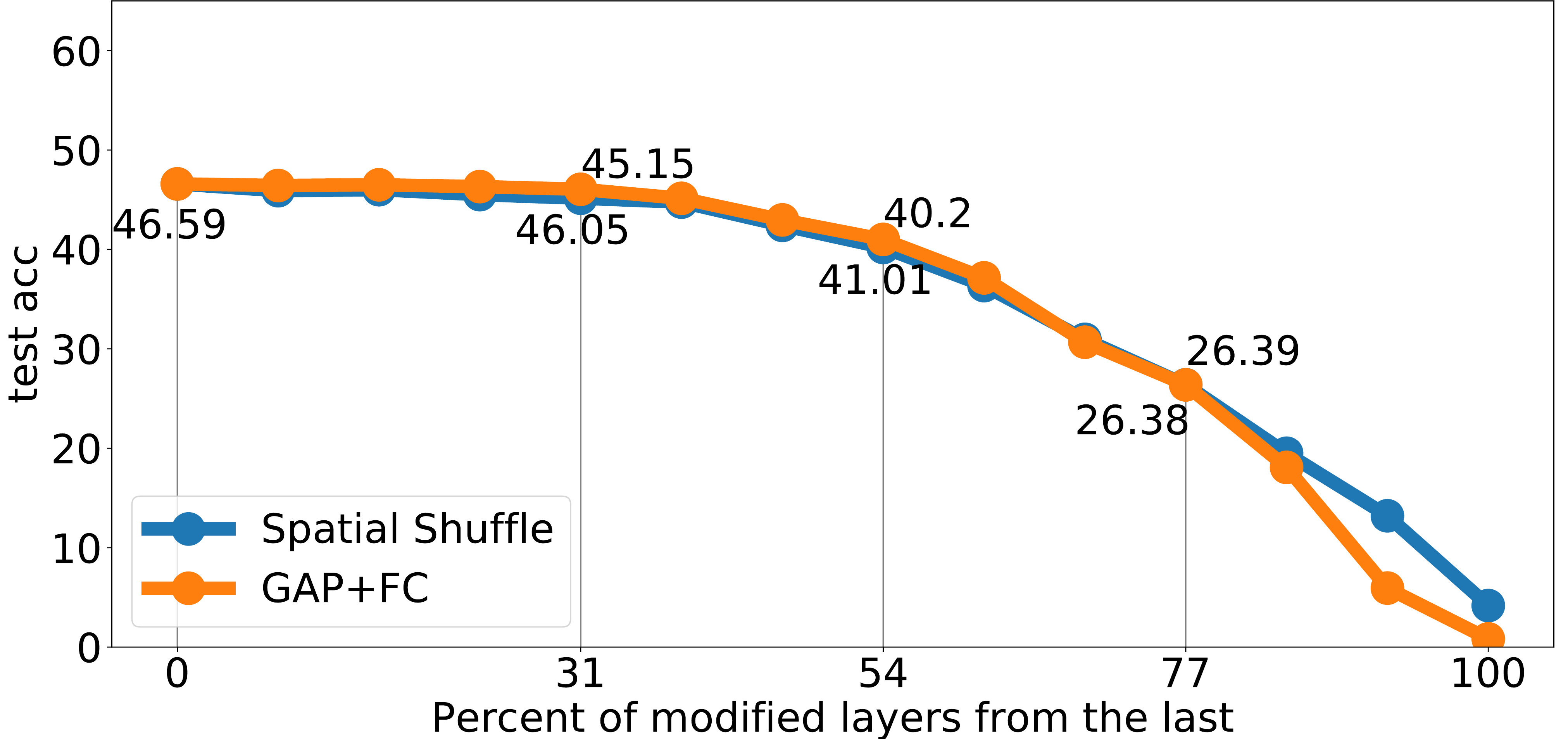}
	\end{minipage} 
	
	\begin{minipage}{0.05\linewidth}
	    \centering
		\includegraphics[scale=0.5,angle=90,origin=c]{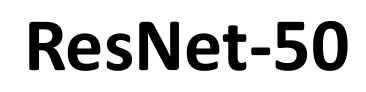}
	\end{minipage} 
	\begin{minipage}{0.45\linewidth}
	    \centering
		\includegraphics[width=1\linewidth]{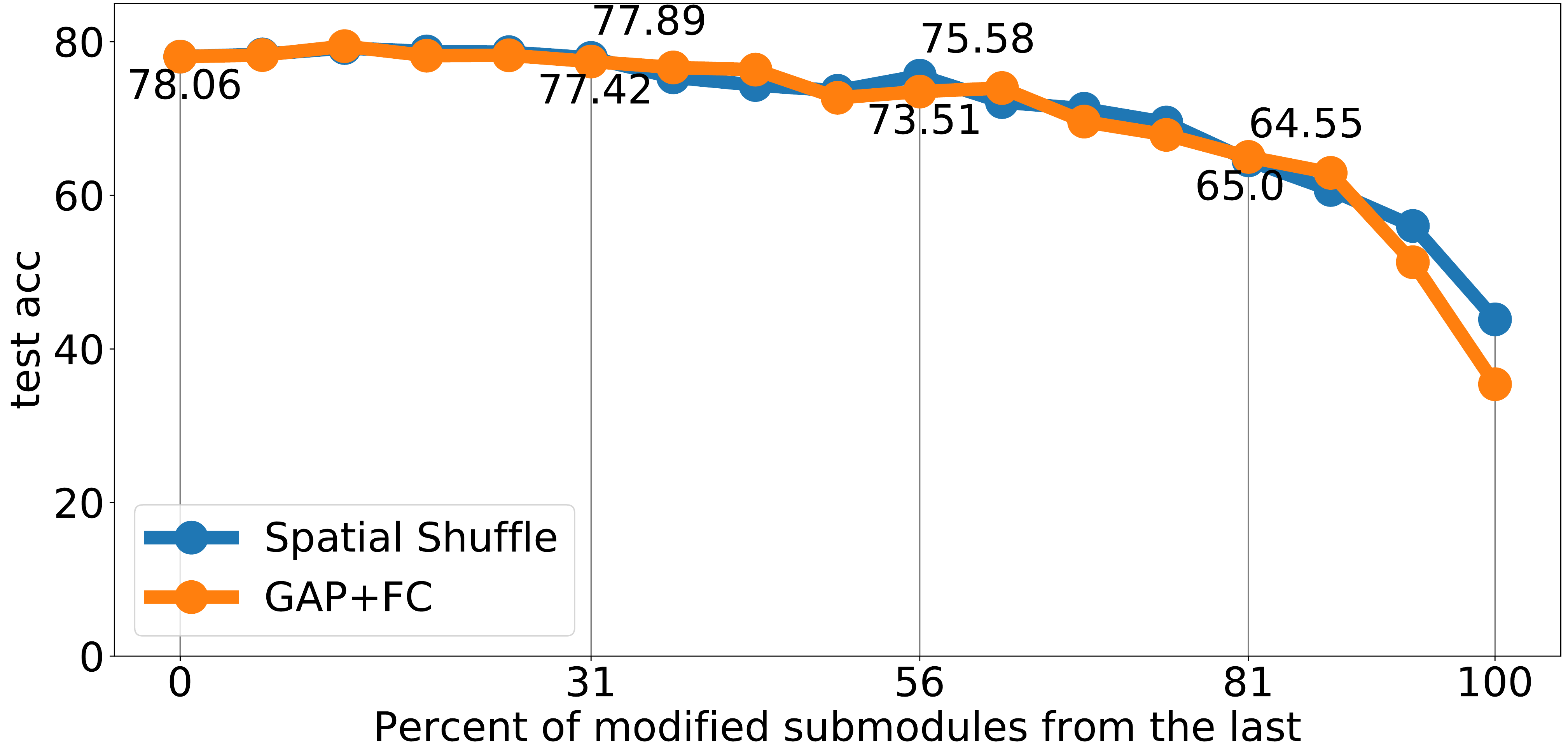}
	\end{minipage} 
	\begin{minipage}{0.45\linewidth}
	    \centering
		\includegraphics[width=1\linewidth]{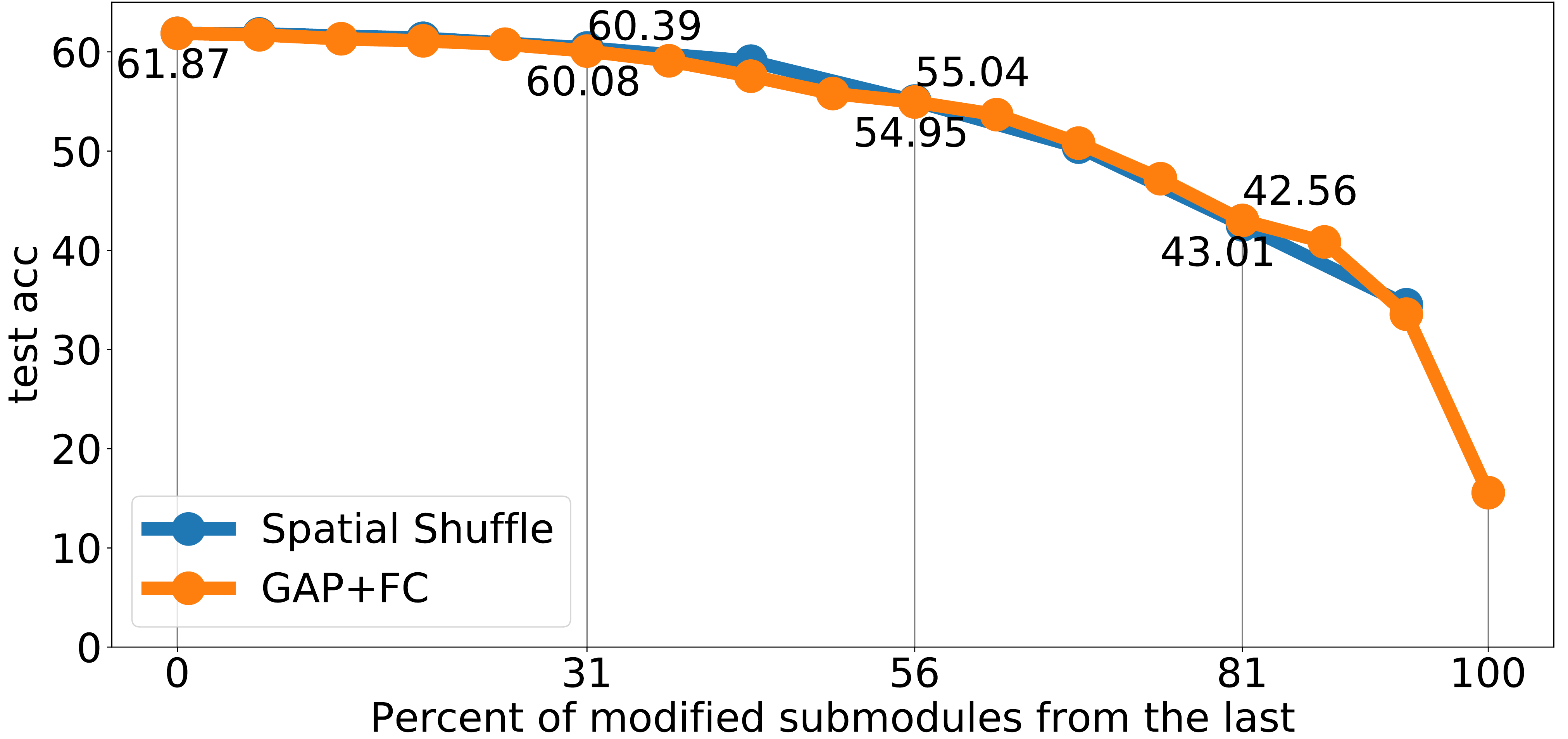}
	\end{minipage} 
	\vspace{7pt}
	\caption{
    Classification results for VGG-16 and ResNet-50 on CIFAR100 and Small-ImageNet-32x32. 
    X-axis is the percent of modified layers / sub-modules counting from the last one. 
    Models on the same dataset are trained with the same setup. 
    It can be observed consistently across experiments that the baseline performance is preserved for a long time despite spatial information is eliminated from last several layers by spatial shuffle or GAP+FC, suggesting that spatial information at later layers is not necessary for a good test accuracy.
    The difference between the baseline models and the models whose latter half of the layers are modifid by GAP+FC or spatial shuffle is, however, still in a reasonable range between 2.48\% (ResNet-50 with spatial shuffle on CIFAR100) to 6.92\% (ResNet-50 with GAP+FC on Small-ImageNet-32x32).
	}
  \label{fig:vgg16_resnet50_cifar100_small}
\end{figure*}
\textbf{Results on CIFAR100 and Small-ImageNet-32x32:}
Results of VGG-16 and ResNet-50 on CIFAR100 and Small-ImageNet-32x32 are shown in Fig. \ref{fig:vgg16_resnet50_cifar100_small}. 
The x-axis is the percent of modified later layers and 0 is the baseline model performance without modifying any layer. 

As we can see, \textit{Spatial Shuffle} and \textit{GAP+FC} have an overall similar behavior consistently across architectures and datasets: the baseline performance is retained for a long time before it starts to decrease with the increase of the percent of modified layers.  
When the last 30\% layers are modified by GAP+FC or spatial shuffle, there are no or little performance decrease across experiments (0.17\% for ResNet-50 on CIFAR100 and 1.44\% for VGG-16 on Small-ImageNet with spatial shuffle).
And the performance decrease is still in a reasonable range (2.48\% with spatial shuffle on CIFAR100 and 6.92\% for GAP+FC on Small-ImageNet-32x32 for ResNet-50) even with around half of the last layers modified.
At 77\% to 81\% of the modified later layers, the performance just starts to show a big difference to the baseline in the range of 8.58\% (ResNet-50 with spatial shuffle on CIFAR100) to 20.21\% (VGG-16 with GAP+FC on Smalll-ImageNet-32x32).


Our experiments here clearly show that spatial information can be neglected from a significant number of later layers with no or small performance drop if the invariance is imposed at training, which suggests that \textit{spatial information at last layers is not necessary for a good performance}.
We should however notice that it does not indicate that models whose prediction is based on spatial information can not generalize well. 
Besides, unlike the common design manner that layers at different depth inside the network are normally treated equally, e.g. the same module is always used throughout the architecture~\cite{howardmobilenets, sandler2018mobilenetv2, iandola2016squeezenet}, our observation implies it is beneficial to have different designs for different layers since there is no necessity to encode spatial information in the later layers. 
As a side effect, GAP+FC can reduce the number of model parameters with little performance drop. 
For example, GAP+FC achieves nearly identical results ($46.05\%$) to the VGG-16 baseline~($46.59\%$), while reducing the number of parameters from $37.70$M to $29.31$M on Small-ImageNet-32x32.

\textbf{Results on ImageNet: }
We further verify our observation on the full ImageNet, where we first reproduce baselines as in the original papers and then apply the same training scheme~\cite{simonyan2014very, he2016deep} directly to train our models. Results are summarized in table~\ref{tab:imgnet}.
We observe that spatial information can be ignored at later layers with a little performance drop on ImageNet as well. 
Considering the complexity of the dataset, the model invariance to the loss of spatial information should intuitively becomes less. 
For example, within 1\% test accuracy drop on ResNet-50, we can manipulate the last 6\% layers by spatial shuffle or GAP+FC.
In ResNet-152, the last 10\% layers can be modified by GAP+FC (the final feature map size is $14 \times 14$) with 0.73\% test accuracy drop. 
However, a simple model as VGG-16 is able to achieve a slightly higher test accuracy than the baseline with spatial shuffle applied to the last 8 \% layers.
The experiments on ImageNet confirm again the previous observation that spatial information at later layers is not necessary for a good performance. Besides, modification from GAP+FC always has a less number of model parameters than that from spatial shuffle with similar test accuracy.

\begin{table}[]
\centering
\small\addtolength{\tabcolsep}{-3pt}
\begin{tabular}{cc|ccc}
Model     & Method   & Top-1(\%) & \begin{tabular}[c]{@{}l@{}}\#Modified\\ Layers(\%)\end{tabular} & \#Params(M) \\ \hline \hline
          & baseline & 72.96     & - & 37.70       \\
VGG16     & spatial shuffle  &   73.51    & 8  & 37.70       \\
          & GAP+FC   & 72.55     & 8 &      35.61       \\ \hline
          & baseline & 75.47     & - & 25.56       \\
ResNet50  & spatial shuffle  & 74.68     & 6 & 25.56       \\
          & GAP+FC   & 74.57     & 6 &      23.46       \\ \hline
          & baseline & 77.66     & - & 60.19       \\
ResNet152 & spatial shuffle  &    76.64   & 2  & 60.19       \\
          & GAP+FC   & 76.93     & 10 &  52.85     
\end{tabular}
\vspace{7pt}
\caption{ImageNet classification results for ResNet-152, ResNet-50 and VGG-16 with spatial shuffle and GAP+FC shows spatial information at later layers is not necessary as well and GAP+FC can reduce the number of parameter with a minor drop of the test accuracy. The best performed models are selected. }
\label{tab:imgnet}
\end{table}

\subsection{Single Layer and Multiple Layers Shuffle} \label{sec:single}
Previous experiments apply random spatial shuffle from one specific layer to the last layer in a network in order to prevent the model from ``memorizing'' encountered permutations. 
Memorization of random patterns is something that deep networks have been shown to be powerful at~\cite{zhang2016understanding}.
We show here the difference between shuffling a single layer and shuffling multiple layers at a time as a sanity check to see whether the model is able to recover the damage to the spatial information done by the random shuffle.

\begin{figure}
	\centering
	\includegraphics[width=0.8\linewidth]{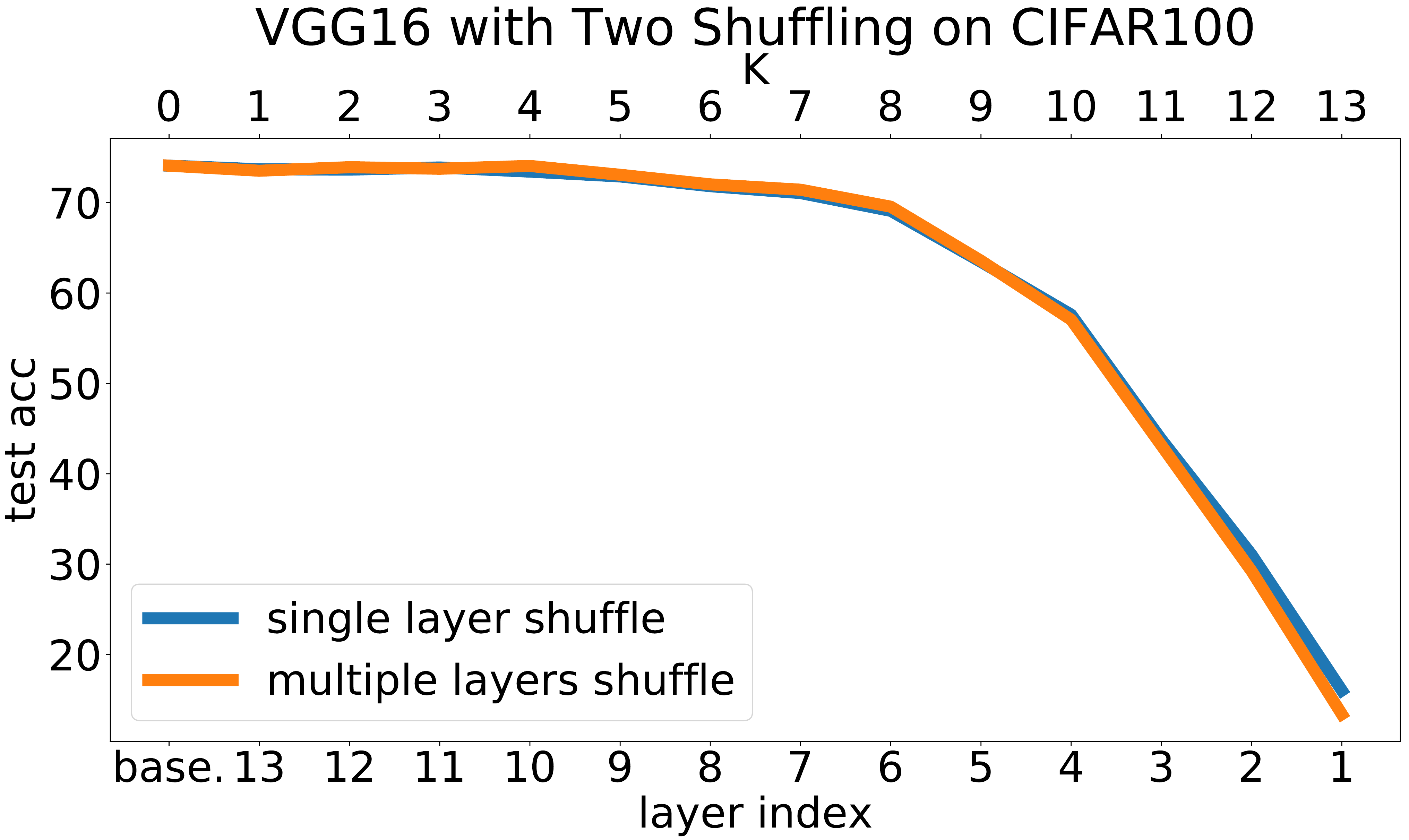}
	\caption{The orange curve is the test accuracy of the multiple layer spatial shuffling with K the number of shuffled layers counting from the last one (0 is the baseline model); the blue curve is the test accuracy of the single-layer spatial shuffling. The x-axis is the layer index with 13 being the last convolutional layer in VGG-16. Random shuffle is applied both at training and test time. The result implies random spatial shuffle at a single layer has very similar effect on the test performance as that at multiple layers.}
	\label{fig:single_shuffle}
\end{figure}

The result of VGG-16 on CIFAR100 is summarized in Fig. \ref{fig:single_shuffle} where the x-axis is the layer index (VGG-16 has 13 convolutional layers) for single layer shuffling and the number of consecutively shuffled layers for multiple layers shuffling.
Highly overlapped curves indicate a similar effect of multiple layer shuffling and single layer shuffle.
We therefore have enough evidence to believe that the model is not able to recover the damage caused by shuffling an early layer.
So in the following experiments, we will be using the single layer shuffle due to the computational burden imposed by shuffling multiple layers.

\subsection{Patch-wise Spatial Shuffle} \label{sec:localshuffle}
In this section, we study the relation between the model performance and the amount of spatial information that can be propagated throughout a network.
The latter is controlled by patch-wise spatial shuffle with different patch sizes.
The larger the patch size is, the less the spatial information is preserved.
Patch-wise spatial shuffle reduces to spatial shuffle when the patch size is the same as the feature map size, in which case no spatial information remains.
Our experiments are conducted on CIFAR100 for VGG-16 and ResNet-50 and and we only shuffle a single layer at a time based on the result in \ref{sec:single}.

\begin{figure}
    \centering
	\begin{minipage}{0.49\linewidth}
		\centering
		\includegraphics[width=1\linewidth]{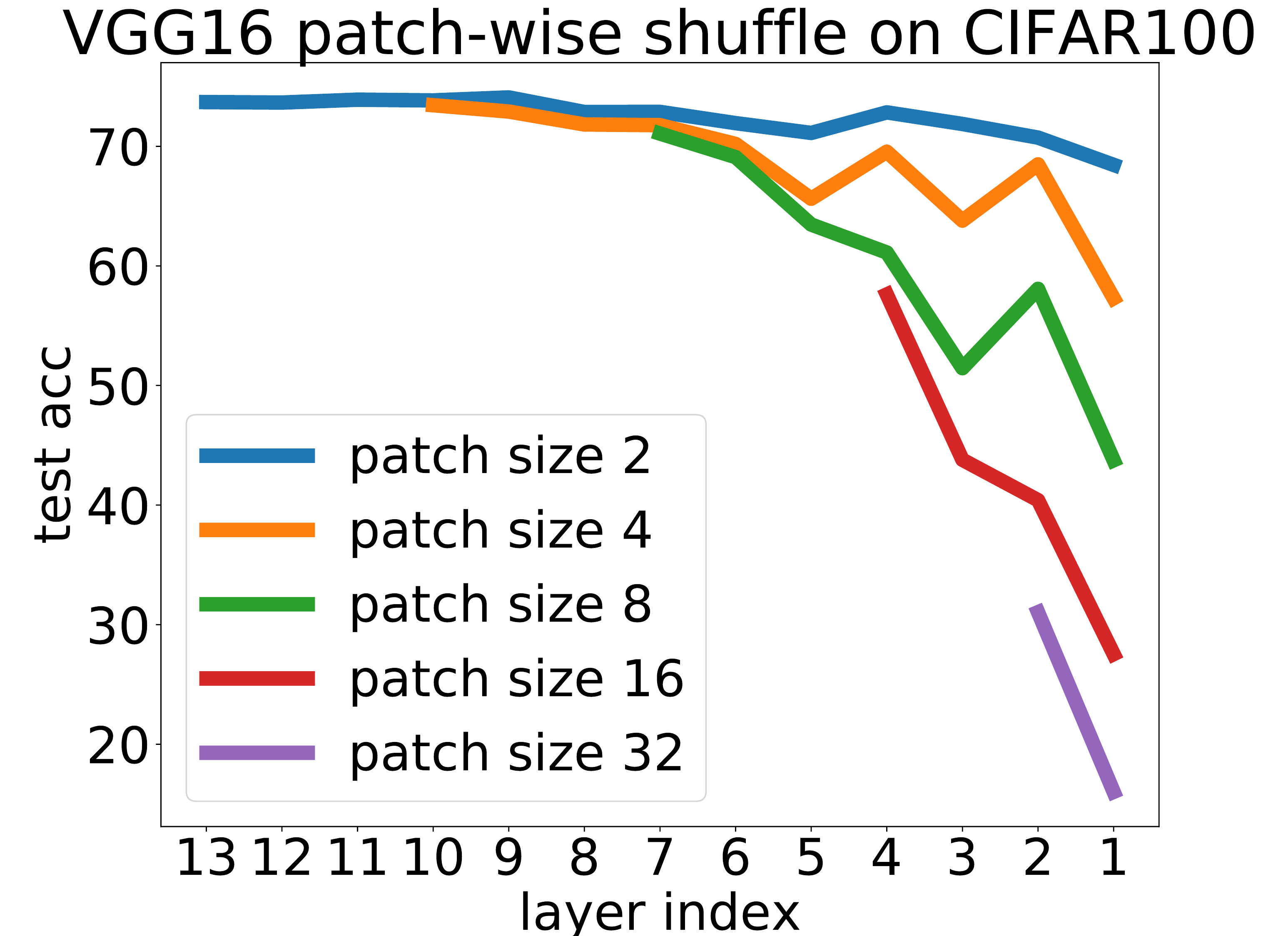}
	\end{minipage} 
	\begin{minipage}{0.49\linewidth}
		\centering
		\includegraphics[width=1\linewidth]{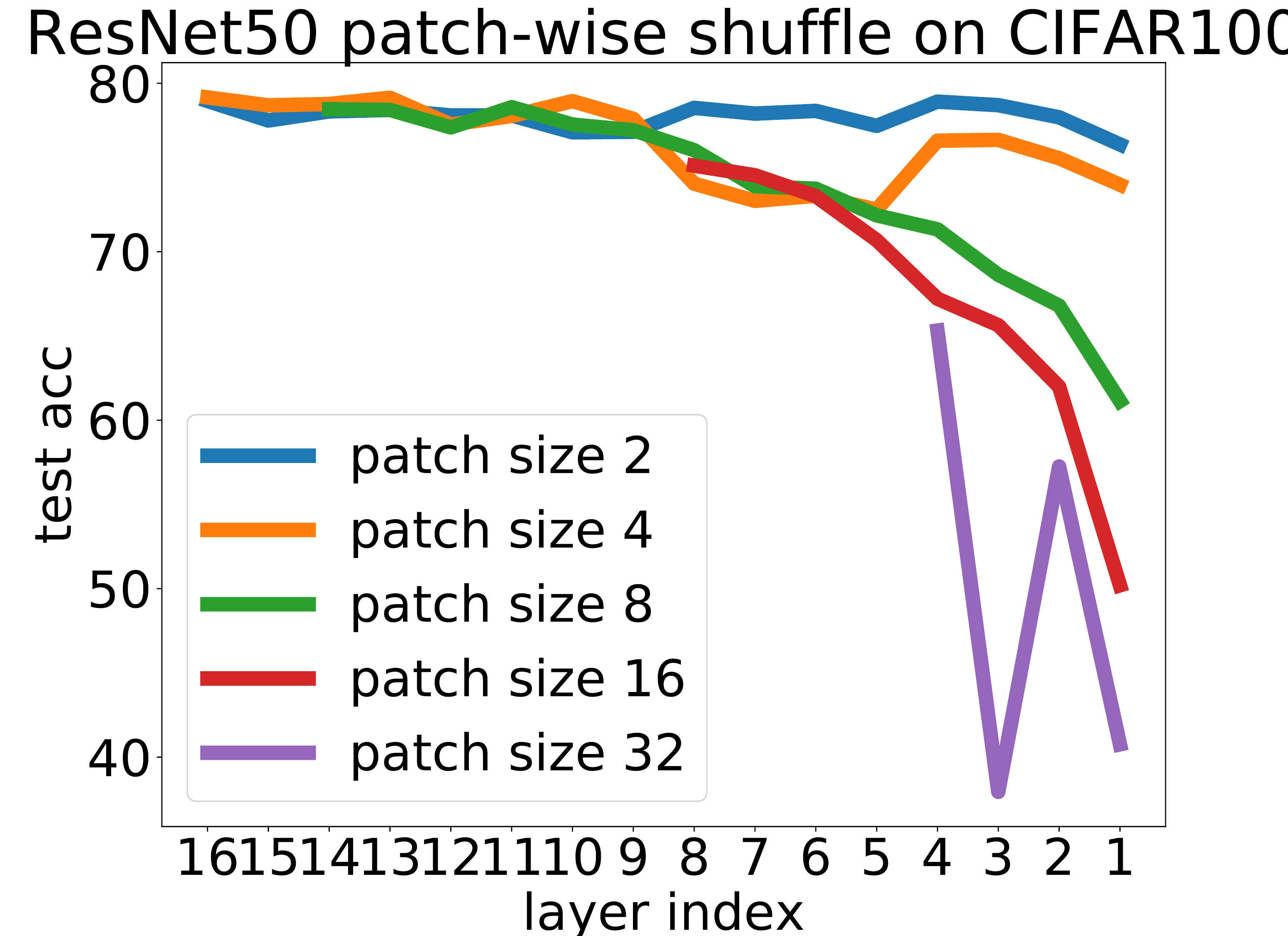}
	\end{minipage} 
	\vspace{7pt}
	\caption{
	The result of patch-wise spatial shuffling of VGG-16 and ResNet-50 on CIFAR100. Only a single layer is shuffled at a time. Layer index 13 and 16 stand for the last layer of VGG-16 and ResNet-50, respectively. With the increase of the patch size, the test accuracy deceases faster at first layers than that at last layers. It is interesting  to see that both models' performance don't fall into random guess (16.02\% for VGG-16 and 40.76\% for ResNet-50) at layer index 1 and patch size 32, where the input image is completely shuffled.
	}
	\label{fig:localshuffle_result}
\end{figure}

The results of patch-wise spatial shuffling of different patch sizes is shown in Fig. \ref{fig:localshuffle_result}.
We can see that the patch size does not make much difference in terms of the test accuracy at later layers, e.g. patch size 2, 4 and 8 for ResNet-50 at 8-14 layers are similar.
However, the performance has a rapid decrease with the increase of the patch size at first layers, indicating a relatively important role of spatial information at first layers.
Nevertheless, this role might not be as much important as what is commonly believed as the ResNet-50 can somehow tolerate a patch-size-4 shuffling on the input image with a small performance drop (4.0\% accuracy difference to the baseline), which is quite impressive given the small size ($32 \times 32$) of the input image of CIFAR100 dataset.

\subsection{Detection Results on VOC Datasets} \label{sec:det}

\begin{figure*}[!ht]
	\centering
	\includegraphics[width=1\linewidth]{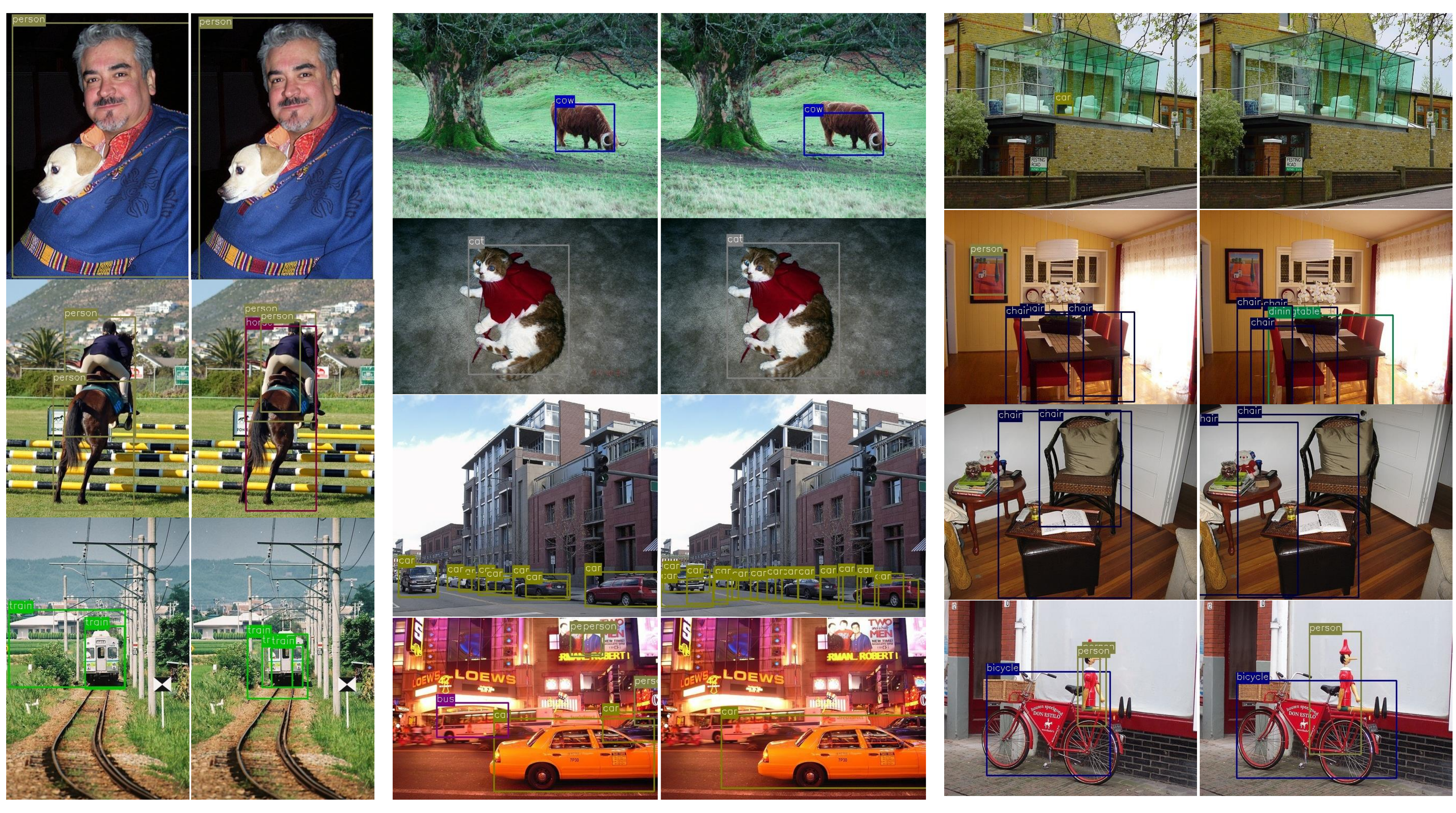}
	\caption{
	Qualitative detection results on the VOC 2007 test set. Examples are the first 11 images in the test set. The left result is from the baseline, and the right result is from the shuffled model. We can see that the shuffled model is not good at small objects, e.g. it missed the people on the ad board in the bottom middle image and it drew too many bbox for the small train and cars. However, its performance on large object seems better than the baseline: it predicted the horse, the dining table and the puppet correctly and precisely.
	}
  \label{fig:det_examples_ref}
\end{figure*}

\begin{figure}[t]
	\centering
	\begin{minipage}{0.49\linewidth}
		\centering
		\includegraphics[width=1\linewidth]{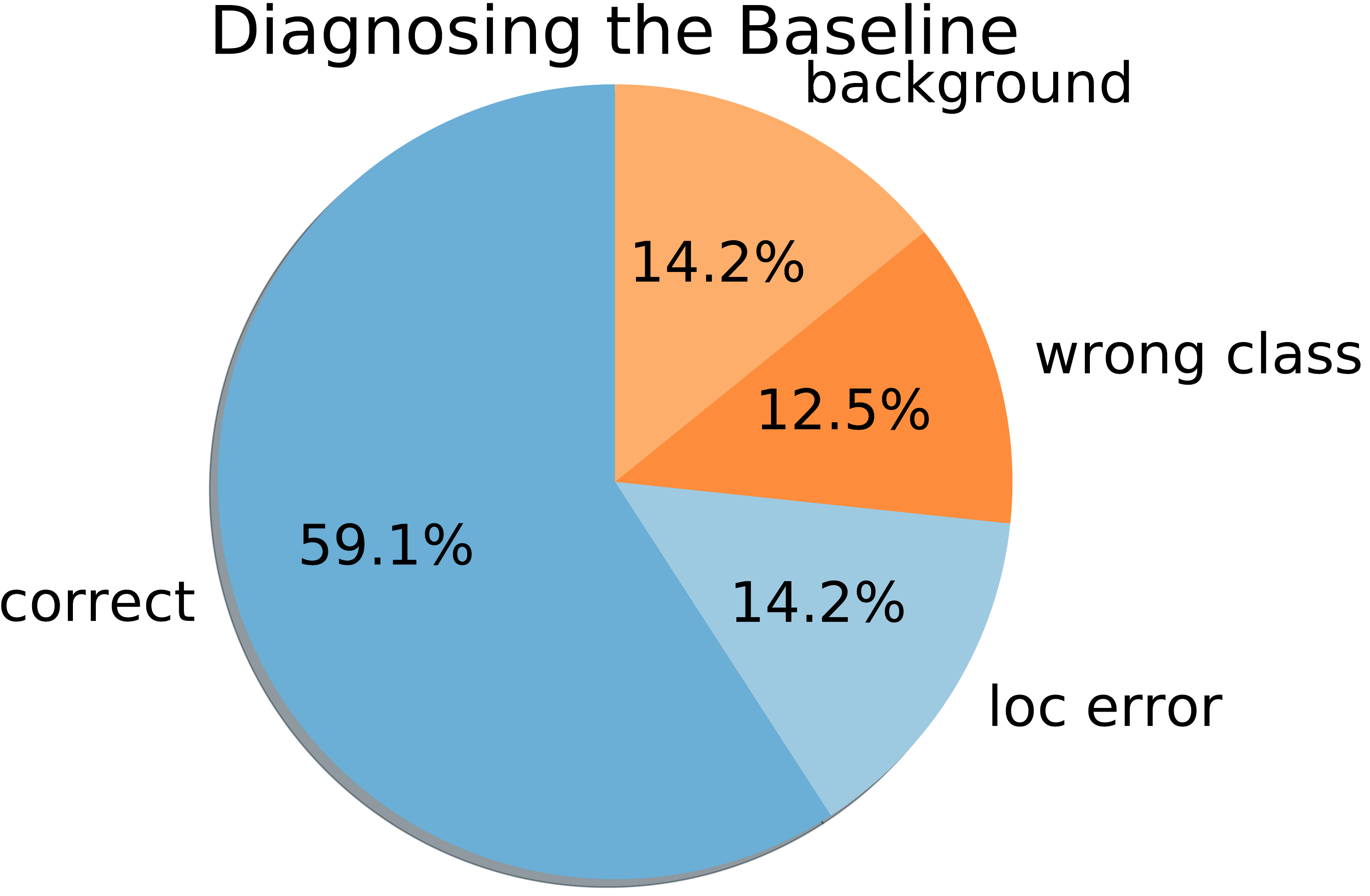}
	\end{minipage} 
	\begin{minipage}{0.49\linewidth}
		\centering
		\includegraphics[width=1\linewidth]{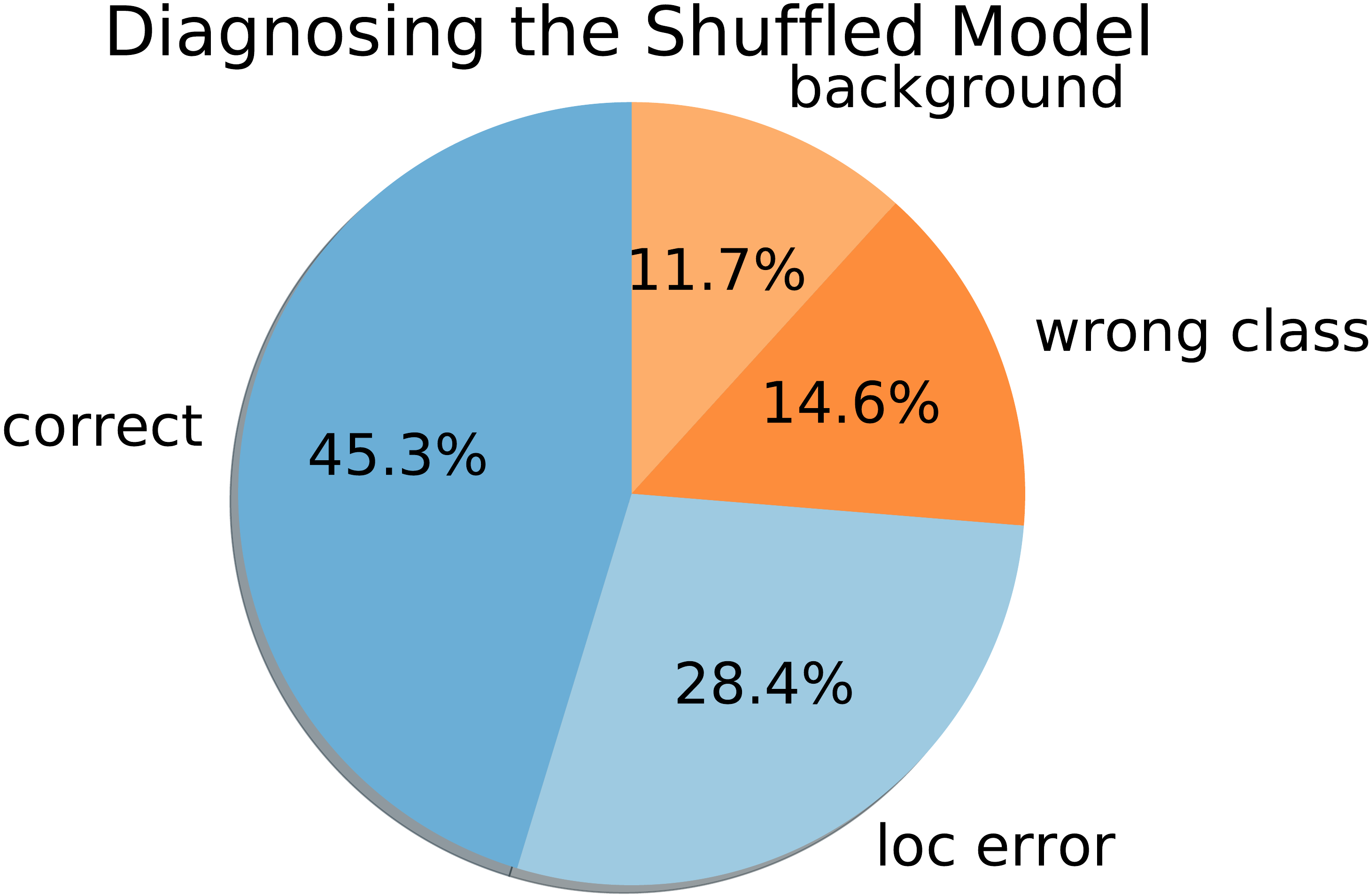}
	\end{minipage} 
	\vspace{7pt}
	\caption{Detection error analysis of our baseline and the shuffled model shows a doubled localization error in the shuffled model and the rest types of error are in the same level with the baseline. 
	}
  \label{fig:det_error}
\end{figure}


Object detection should intuitively suffer more from spatial shuffling than classification since the spatial information should help localizing objects.
In this section, we show some initial results on Pascal VOC \cite{voc2012, voc2007}.

We design an analogue to YOLO~\cite{redmon2016yolo} as our detection model. 
The architecture consists of a backbone and a detection head; the backbone is a ResNet-50 without the classifier and the detection head has 3 bottlenecks and a $3 \times 3$ convolutional layer whose outputs is in the same format as \cite{redmon2016yolo}. 
Different to \cite{redmon2016yolo}, we deploy a $3 \times 3$ convolution instead of a fully connected layer in the end to output the final detection results.
The latter gives the model potential access to the object feature which may be exploited by the model to predict its location.
In order to prevent the undesirable shortcut, we use a $3 \times 3$ convolution so that the prediction of a bounding box at a certain location does not depend on all activation on the feature map.

By using a pre-trained ResNet-50 on ImageNet, we are able to reach 66\% mAP on VOC2007 test set after fine-tuning, which is the same as the number in \cite{redmon2016yolo}.
To avoid pretraining a spatially shuffeled model on ImageNet, we compare a spatially shuffled
model and a non spatially shuffled model, both trained from scratch on VOC.
Our models are trained for 500 epochs with exponentially decaying learning rate starting from 0.001. Our baseline model achieves 50\% mAP on VOC2007 test set without using an ImageNet pre-trained backbone.
The result of the shuffled model, where we apply random shuffle to the last layer of the backbone, is 34\%. 
While this sounds like a large drop it turns out that the classification performance is essentially preserved and 
only the localization performance is suffering. To analyze this effect in detail, we use the method and tools proposed in \cite{diag12hoiem}.
The diagnosis tool classifies each prediction from the model as either correct prediction or a type of error based on its class label and IoU with the ground truth. More details about the diagnosis method can be found in \cite{diag12hoiem}.


The result in Fig. \ref{fig:det_error} shows that the mis-classification to the wrong class and background are of similar percents for both models, and the localization error is doubled for the shuffled model (increase from 14.2\% to 28.4\%).
Though random shuffling indeed affects the model's localization ability, it is unexpected that the effect is not fatal given that it is highly likely the model trained with spatial shuffle has to predict the correct bounding box for one object based on some other features since random shuffling switches features.
We should also notice that a prediction is counted as a localization error if it has the correct class label and the IoU to the ground truth is less than 0.5.
Therefore, classification-wise speaking, the shuffled model got 73.7\% (45.3\% + 28.4\%) of its predictions correct, which is even slightly higher than for the baseline (73.3\% = 59.1\% + 14.2\%).

\textbf{Qualitative Results: }
Fig. \ref{fig:det_examples_ref} shows some qualitative results from both models. 
Those examples are the first 11 images in the VOC2007 test set.
We can see that the localization error actually mainly comes from small objects for which the shuffled model tends to predict several bounding boxes on one object, and the bounding box of the relatively big object is not really off, e.g. the shuffled model managed to localize the dining table in the middle right image and the horse in the middle left image while the baseline can not.

\section{Conclusion}
To conclude, we empirically show that a significant number of later layers of CNNs are robust to the absence of the spatial information, which is commonly assumed to be important for object recognition tasks. 
Modern CNNs are able to tolerate the loss of spatial information from the last 30\% of layers at around 1\% accuracy drop; and the test accuracy only decreases by less than 7\% when spatial information is removed from the last half of layers on CIFAR100 and Small-ImageNet-32x32.
Though depth of the network is essential for good performance, the later layers do not necessarily have to be convolutions.


{\small
\bibliographystyle{ieee_fullname}
\bibliography{egbib}
}

\end{document}


\title{Supplemental Material \\}

\author{First Author\\
Institution1\\
Institution1 address\\
{\tt\small firstauthor@i1.org}
\and
Second Author\\
Institution2\\
First line of institution2 address\\
{\tt\small secondauthor@i2.org}
}

\maketitle

\section{GAP+FC Outperforms the Baseline without Data Augmentation}
Another interesting point is that GAP+FC have a regularization effect that can lead to improved generalization performance when data augmentation is not applied.
Table \ref{tab:da_resnet50_cifar100} shows the test accuracy of modified ResNet-50 via GAP+FC trained with and without data augmentation on CIFAR100. 
While the models trained with data augmentation show similar test accuracy, we observe a significant performance improvement over the baseline on ResNet-50 trained without data augmentation, e.g GAP+FC outperforms the baseline by $7.36\%$ with the last 56\% layers replaced by fully-connected layers.

\section{ResNet50 Architecture} \label{app:resnet50_arch}
Fig. \ref{fig:arch_resnet50} shows an example of ResNet-50 with the last 3 bottlenecks being modified by shuffle conv or GAP+FC. Our modification is applied only on the $3 \times 3$ convolution inside each bottleneck since it is the only operation that has spatial extent.
\section{More Detection Results}
\begin{figure*}[!ht]
	\centering
    \includegraphics[width=1\linewidth]{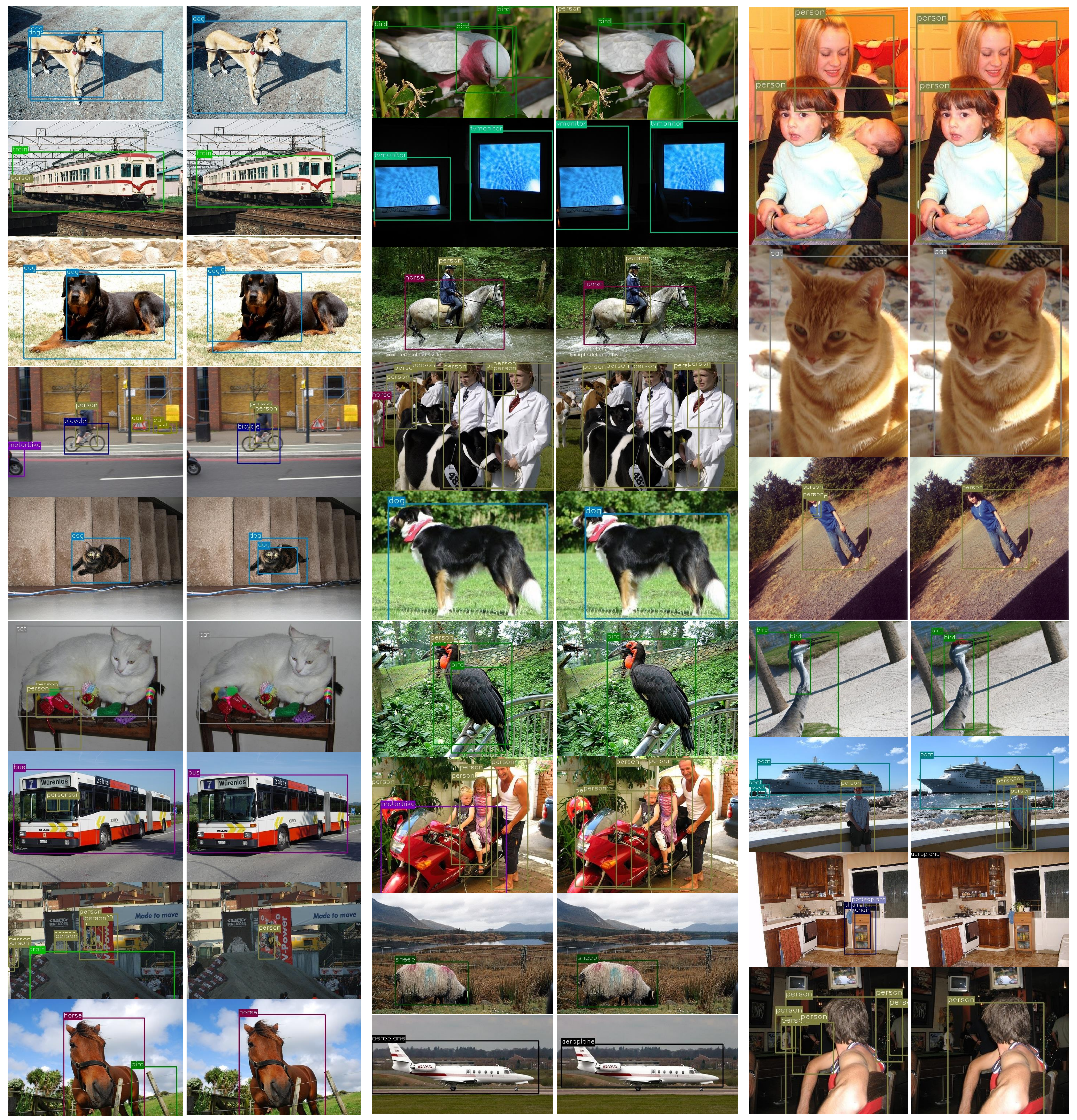}
	\caption{More detection examples on the VOC 2007 test set. The left result is from the baseline, and the right result is from the shuffled model. Each image was sampled randomly. We can see that the shuffled model is not good at small objects, and the localization is not as accurate as the baseline model.}
	\label{fig:det_imgs1}
\end{figure*}

More qualitative results are presented in Fig. \ref{fig:det_imgs1}.
Those images are not handpicked, but randomly selected from the test set. 
The left result is from the baseline, and the right result is from the shuffled model. 
We can see that the shuffled model tends to have a fewer number of responds to small objects than the baseline model, even though not all responds from the baseline model are correct. 
However, the shuffled model shows a reasonable performance on localizing big objects, and sometimes, is even better than the baseline model with less number of repetitive bboxes on the same object.
\begin{table}[]
\centering
\begin{tabular}{cc|cc}
\multicolumn{2}{c|}{Config. for ResNet-50} & Top-1(\%) & \begin{tabular}[c]{@{}l@{}}\#Modified\\ Layers(\%)\end{tabular} \\ \hline \hline
\multirow{2}{*}{w/ DataAug}  & baseline & 78.06 & - \\
                             & GAP+FC   & 79.42 & 12\% \\ \hline
\multirow{2}{*}{w/o DataAug} & baseline & 65.64 & - \\
                             & GAP+FC   & 73.00 & 56\% 
\end{tabular}
\caption{Effect of data augmentation on classification results for ResNet-50 on CIFAR100. The data augmentation here is the random flipping and the random cropping. We present here the best performed model for GAP+FC. We can see that modified models reach much higher test accuracy when data augmentation is not applied. ResNet-50 with GAP+FC trained without data augmentation shows a significant performance improvement over the baseline from 65.64\% to 73.00\% on CIFAR100.}
\label{tab:da_resnet50_cifar100}
\end{table}
\begin{figure}
	\centering
	\includegraphics[width=0.9\linewidth]{images/arch_resnet50_crop.pdf}
	\caption{An example of ResNet-50 with the last 3 bottlenecks being modified by shuffle conv or GAP+FC.}
	\label{fig:arch_resnet50}
\end{figure}
